\documentclass[letterpaper, 10 pt, conference]{ieeeconf}  
\IEEEoverridecommandlockouts
\usepackage{graphicx}
\usepackage{amssymb,amsmath,physics}
\usepackage{array,booktabs,arydshln}
\usepackage[cal=boondox,scr=boondoxo]{mathalfa}
\usepackage{algorithm}
\usepackage{algorithmic}
\usepackage[utf8]{inputenc}
\usepackage[english]{babel}
\usepackage{xcolor}
\definecolor{green}{RGB}{3,112,15}
\definecolor{yellow}{RGB}{255,140,0}
\usepackage{hyperref}
\usepackage{cite}
\usepackage{bm}
\usepackage{sidecap}
\usepackage{subcaption}

\newcommand{\dynalflh}{Dyna-LfLH}

\DeclareMathOperator*{\argmin}{argmin}

\title{\large \bf
Dyna-LfLH: Learning Agile Navigation in Dynamic Environments from Learned Hallucination
}

\author{Saad Abdul Ghani$^{1}$, Zizhao Wang$^{2}$, Peter Stone$^{2, 3}$, and Xuesu Xiao$^{1}$
\thanks{
$^{1}$George Mason University {\tt\small \{sghani2, xiao\}@gmu.edu}
$^{2}$The University of Texas at Austin {\tt\small zizhao.wang@utexas.edu, pstone@cs.utexas.edu}
$^{3}$Sony AI
}
\thanks{This work has taken place in the RobotiXX Laboratory at George Mason University and the Learning Agents Research Group (LARG) at the Artificial Intelligence Laboratory, The University of Texas at Austin. RobotiXX research is supported by National Science Foundation (NSF, 2350352), Army Research Office (ARO, W911NF2320004, W911NF2420027, W911NF2520011), Air Force Research Laboratory, US Air Forces Central, Google DeepMind, Clearpath Robotics, Raytheon Technologies, Tangenta, Mason Innovation Exchange, and Walmart. LARG research is supported in part by NSF (FAIN-2019844, NRT-2125858), Office of Naval Research (N00014-24-1-2550), ARO (FAIN W911NF-17-2-0181, W911NF-23-2-0004, W911NF-25-1-0065), DARPA (Cooperative Agreement HR00112520004 on Ad Hoc Teamwork), Lockheed Martin, and Good Systems, a research grand challenge at the University of Texas at Austin. The views and conclusions contained in this document are those of the authors alone. Peter Stone serves as the Chief Scientist of Sony AI and receives financial compensation for that role. The terms of this arrangement have been reviewed and approved by the University of Texas at Austin in accordance with its policy on objectivity in research.}
}

\begin{document}

\maketitle
\begin{abstract}
This paper introduces Dynamic Learning from Learned Hallucination (\dynalflh), a self-supervised method for training motion planners to navigate environments with dense and dynamic obstacles. Classical planners struggle with dense, unpredictable obstacles due to limited computation, while learning-based planners face challenges in acquiring high-quality demonstrations for imitation learning or dealing with exploration inefficiencies in reinforcement learning. Building on Learning from Hallucination (LfH), which synthesizes training data from past successful navigation experiences in simpler environments, Dyna-LfLH incorporates dynamic obstacles by generating them through a learned latent distribution. This enables efficient and safe motion planner training. We evaluate Dyna-LfLH on a ground robot in both simulated and real environments, achieving up to a 25\% improvement in success rate compared to baselines.

\end{abstract}

\section{Introduction}
\label{sec::intro}

Dynamic obstacles present a significant challenge for autonomous mobile robots, requiring them to adapt their motion plans in real-time to avoid collisions. Pedestrians crossing streets unexpectedly or other robots performing independent tasks in warehouses exemplify the types of dynamic obstacles that challenge autonomous mobile robots. Such obstacles are often characterized by unpredictable motion patterns, demanding intelligent navigation strategies that can anticipate and respond to rapid changes in the environment.

Navigating around unpredictable dynamic obstacles in a highly dense environment poses significant computational challenges for classical navigation systems, making it inefficient to draw samples or perform optimization iterations in real-time. Recently, machine learning approaches have been used to successfully maneuver around such obstacles in a data-driven manner~\cite{xiao2022motion, wang2021agile}. However, both Imitation Learning (IL) and Reinforcement Learning (RL) approaches depend on high-quality training data that is difficult and inefficient to acquire in the former and latter cases respectively.

Learning from Hallucination (LfH)~\cite{xiao2021toward, xiao2021agile, wang2021agile, park2023learning} is a paradigm that can safely and efficiently provide a variety of training data for collision avoidance without the need of actually training in challenging obstacle configurations. In LfH, the robot gathers motion plans from past navigation experiences in relatively easy or completely open environments, imagines other more difficult obstacle configurations for which the existing motion plans would also be optimal (i.e., hallucination), and then learns a motion planner based on the hallucinated obstacle configurations and motion plans. This process circumvents the data dependency of IL and RL as one can generate a large amount of data safely and efficiently without the need for an expert supervisor or trial-and-error exploration. However, existing LfH methods are only designed to hallucinate static environments and fail to perform well in dynamic ones. 

\begin{figure}
  \centering
  \includegraphics[width=0.8\columnwidth]{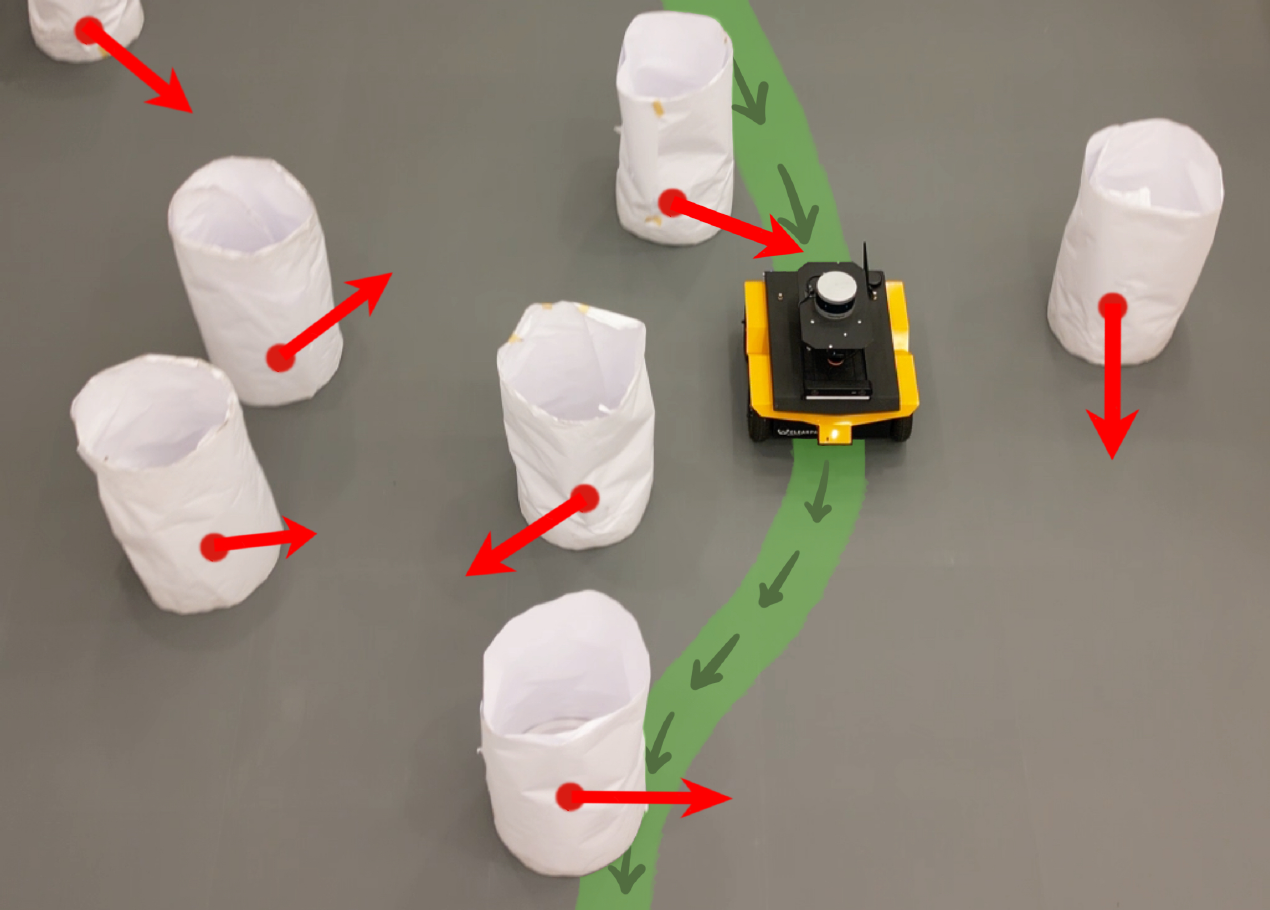}
  \caption{A mobile robot navigating through a dense and dynamic obstacle field using \dynalflh. }
  \label{fig::dynaInAction}
  \vspace{-10pt}
\end{figure}

In this paper, we propose a new Dynamic Learning from Learned Hallucination (\dynalflh) approach (Fig.~\ref{fig::dynaInAction}). We design a novel latent distribution that can be learned through \dynalflh\ in a self-supervised manner and then sampled from to generate a variety of dynamic obstacle configurations. Paired with existing optimal motion plans, these dynamic obstacle configurations are used to learn a motion planner to navigate in environments filled with a large number of dynamic obstacles. \dynalflh\ is tested on a ground robot both in simulated and physical environments. Superior navigation performance is achieved when compared to LfLH~\cite{wang2021agile}, a classical sampling-based motion planner (DWA)~\cite{fox1997dynamic}, a state of the art sampling-based model predictive controller (Log-MPPI)~\cite{log_mppi}, and an IL method~\cite{amir2023socialnav}.

\section{Related Work}
\label{sec::related}
This section reviews classical motion planning and recent machine learning techniques for mobile robot navigation in dynamic environments. We also introduce the recent LfH paradigm, which our \dynalflh\ belongs to. 

\subsection{Classical Motion Planning}
Two popular approaches~\cite{mohanan2018survey} to solve motion planning in dynamic environments are Artificial Potential Fields (APF)~\cite{quinlan1993elastic} and velocity-based methods~\cite{fox1997dynamic}. In APF, the environment is modeled as a field of attractive and repulsive forces, guiding the robot through space. The goal has an attractive potential field while obstacles have a negative potential field. The resultant force is calculated to guide the robot towards the goal and away from the obstacles. 
Velocity-based methods directly work on the robot's and obstacles' kinematics and dynamics. First, the robot's and obstacles' kinematics are taken into account and an initial kinematic trajectory is created to avoid the obstacles. Then, using the robot's dynamics, a motion plan is created to closely follow the initial kinematic trajectory. Dynamic Window Approach (DWA)~\cite{fox1997dynamic} is a well-known example of a velocity-based method.

Compared to APF and velocity-based motion planners, \dynalflh\ uses configuration space, which decomposes the environment into free and obstacle spaces. An advantage of our learning approach is that its computation is not dependent on obstacle density and movement during deployment, because \dynalflh\ simply queries a pre-trained neural network to produce feasible and fast navigation behaviors.

\subsection{Machine Learning for Navigation}
Machine learning approaches have been applied to mobile robot navigation in different ways~\cite{xiao2022motion}, such as either applying learning in conjunction with classical methods~\cite{xiao2022appl} or using IL~\cite{xiao2022learning} or RL~\cite{tai2017virtual} to learn an end-to-end planner~\cite{wullt2023neural}.   

Most learning methods require either high-quality (IL)  or extensive (RL) training data. \dynalflh\ is a self-supervised learning approach that automatically generates diverse training data, addressing the conundrum of needing to know what good navigation behavior is, without prior knowledge of how to achieve it.

\subsection{Learning from Hallucination (LfH)}
LfH~\cite{xiao2021toward, xiao2021agile, wang2021agile, park2023learning} generates training data by hallucinating obstacle configurations where existing plans are optimal, thus eliminating the need for expert demonstrations or risky exploration. Researchers have designed hallucination techniques to project the \emph{most constrained}~\cite{xiao2021toward}, a \emph{minimal}~\cite{xiao2021agile}, or a \emph{learned}~\cite{wang2021agile} obstacle configuration onto the robot perception. Hallucination has also been used to enable multi-robot navigation in narrow hallways~\cite{park2023learning} and to augment existing global motion plans for which the global path is optimal for~\cite{das2024motion}.

However, all existing LfH approaches assume that the environment is static, or enforce static perception through hallucinating virtual fields~\cite{park2023learning}. In this work, we generalize the existing LfH formulation into dynamic environments and show our new \dynalflh\ can hallucinate appropriate dynamic obstacles to safely and efficiently provide training data to learn an agile motion planner to navigate through highly-cluttered, fast-moving, hard-to-predict obstacles.

\section{Approach}
\label{sec::approach}

\begin{figure*}
   \centering
   \includegraphics[width=2.0\columnwidth]{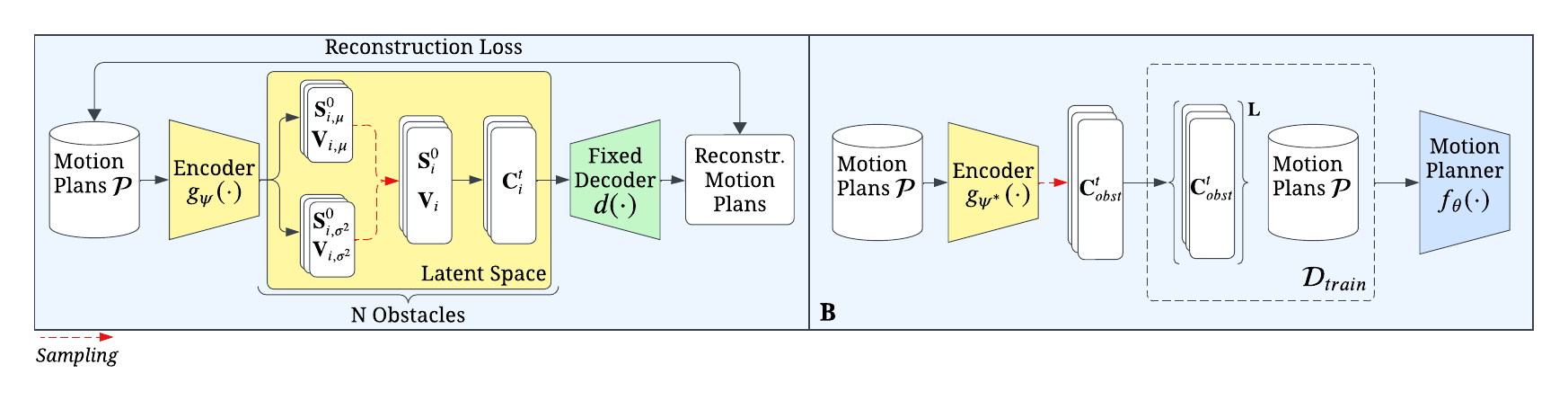}

\caption{\small
\textbf{A.} The encoder-decoder architecture learns the hallucination function $g_{\psi}$ (yellow) in a self-supervised manner using past motion plans. The latent space consists of vectors $\mathbf{S}^0$ and $\mathbf{V}$, the N hallucinated obstacles' initial locations and velocities which are sampled from a normal distribution with learned parameters ($\mu$ and $\sigma^2$). Using $\mathbf{S}^0$ and $\mathbf{V}$, the obstacles $C^t$ are constructed and passed to the fixed, differentiable decoder $d(\cdot)$ (green). $d(\cdot)$ reconstructs a motion plan, $\hat{p}$, that is optimal given $C^t$. Then, $\hat{p}$ is compared against the original motion plans, $p$. \textbf{B.} Once the hallucination function is trained, we hallucinate and sample S$\times$N dynamic obstacles from $g_{\psi^*}$ that is used to render and create our supervised training set $\mathcal{D}_{train}$. Finally, we train a motion planner $f_\theta(\cdot)$ (dark blue) using a history of $L$ rendered LiDAR scans and our original motion plans, $p$.}
   \label{fig::diagram}
   \vspace{-16pt}
 \end{figure*}

In this section, we introduce our Dyna-LfLH approach. We first formalize the problem of motion planning in dynamic environments and then reformulate its ``inverse'' problem, i.e., dynamic obstacle hallucination using the LfH paradigm.
Finally, we propose an algorithm to learn a dynamic hallucination function from which we can generate a variety of dynamic obstacle configurations to train a motion planner.

\subsection{Problem Definition}
Robot motion planning in static environments is typically formulated in configuration space (C-space), representing the set of all possible robot configurations. In a particular environment, $C$ is partitioned into $C = C_{free} \cup C_{obst}$, where $C_{free}$ denotes collision-free configurations and $C_{obst}$ represents configurations blocked by obstacles or constraints. A motion plan $p \in \mathcal{P}$ consists of actions $p = \{u^t\}_{t=0}^{T-1},~u^t \in \mathcal{U}$, where $\mathcal{P}$ is the plan space over discrete time horizon $T$ and $\mathcal{U}$ is its action space. The motion planning problem is then to find $p = f(C_{obst}|c_c, c_g)$ to move the robot from its current configuration $c_c$, through a sequence of configurations $c^t$, to its goal configuration $c_g$, such that $c^t \cap C_{obst} = \emptyset,~\forall t$ and $p$ is the optimal path from $c_c$ to $c_g$, according to some metric.

To generalize into dynamic environments, the partition becomes time-dependent, i.e., $C = C_{free}^t \cup C_{obst}^t, ~t \in [1,T]$. The new motion planning problem then becomes $p = f(\{C^t_{obst}\}_{t=1}^T|c_c, c_g)$ such that $c^t \cap C_{obst}^t = \emptyset, ~\forall~ t$. For simplicity, we assume the obstacle states ${C^t_{obst}}$ are known during planning, though this assumption is relaxed in our implementation. 

In previous LfH approaches~\cite{xiao2021toward, xiao2021agile, wang2021agile}, the ``inverse'' problem of motion planning, i.e., the hallucination of obstacle configuration space, such that $p$ is optimal, is defined as $\{C^i_{obst}\}_{i=1}^\infty = f^{-1}(p | c_c, c_g)$. Notice that the inverse function is a mapping from a motion plan $p$ to a \emph{set} of obstacle configurations, since multiple obstacle configurations can make $p$ optimal. In most cases, this set is infinitely large. 

In Dyna-LfLH, we generalize the previous static hallucination to a dynamic one, i.e., $\{\{C^{t, i}_{obst}\}_{t=1}^T\}_{i=1}^\infty =\! f^{-1}(p | c_c, c_g)$, that is, generating all possible obstacle configuration \emph{sequences} which make the given motion plan $p$ optimal over the time horizon $T$. 
Since it is impossible to produce all (infinite) possible obstacle sequences, we approximate $\{\{C^{t, i}_{obst}\}_{t=1}^T\}_{i=1}^\infty$ using a learned distribution, from which we can numerously sample a large number of obstacle sequences over time horizon $T$. To be specific, we learn a hallucination function $g$, which outputs such a distribution: 
\begin{equation}
    \{C^t_{obst}\}_{t=1}^T \sim g(p | c_c, c_g).
    \label{eqn::hallucination}
\end{equation}

\subsection{Approximating $\{C^t_{obst}\}_{t=1}^T$ with Discrete Obstacles} 
Despite the enormous space of all possible dynamic obstacle sequences $\{C^t_{obst}\}_{t=1}^T$, it is reasonable to assume that they are composed of a few discrete moving obstacles. Therefore, in this paper, we assume $\{C^t_{obst}\}_{t=1}^T$ can be approximated by $N$ circular moving obstacles $\{O_i\}_{i=1}^N$ with a fixed radius $R$ at coordinate $\mathbf{C}^t_i = (x^t_i, y^t_i)$ moving in a continuous fashion (obstacles cannot teleport) following first-order dynamics:
\begin{equation}
    \mathbf{C}^t_i = \mathbf{S}^0_i + \mathbf{V}_i\cdot t, ~ 1 \leq t \leq T,
    \label{eqn::obstacle}
\end{equation}
where $\mathbf{S}^0_i=(x^0_i, y^0_i)$ is the starting coordinate of obstacle $O_i$ at $t=0$, and $\mathbf{V}_i=(v^x_i, v^y_i)$ is its fixed velocity. 
These assumptions efficiently decompose our original problem of hallucinating the vast space  of obstacle configuration sequences, $\{C^t_{obst}\}_{t=1}^T$, into learning a set of structured parameter distributions of $(x^0_i, y^0_i)$, and $(v^x_i, v^y_i)$, sampling from them and forming dynamic obstacles: 
\begin{equation}
    \{(x^0_i, y^0_i), (v^x_i, v^y_i)\}_{i=1}^N \sim g(p|c_c,c_g). 
    \label{eqn::distribution}
\end{equation}
In this work we assume the obstacles only follow first-order dynamics. In future work, it is easy to add complex higher-order dynamics by learning the distributions of acceleration, jerk, snap, and so on. Though for a short time horizon, modeling obstacles to have a constant velocity is sufficient. 

\subsection{Learning a Parameterized Hallucination Function}
We instantiate the hallucination function $g$ with learnable parameters $\psi$ and learn $g_\psi$ in a self-supervised reconstructive manner using an encoder-decoder structure similar to the static LfLH approach~\cite{wang2021agile}. 
The encoder $g_\psi(p|c_c,c_g)$ takes the current configuration $c_c$, goal configuration $c_g$, and the corresponding plan $p$ as input and produces the probability distributions of $x^0_i$, $y^0_i$, $v^x_i$, and $v^y_i$ for all $N$ dynamic obstacles, as shown in Eqn.~\eqref{eqn::distribution}. We assume $x^0_i$, $y^0_i$, $v^x_i$, and $v^y_i$ are all independent, normally distributed random variables. Then, a potential $\{C^{t}_{obst}\}_{t=1}^T$ is constructed by applying Eqn.~\eqref{eqn::obstacle} on all $N$ dynamic obstacle parameters sampled from the learned distribution.

The decoder is a 2D classical motion planner without any learnable parameters that is used to generate optimal motion plans $\hat{p}$ from the sampled obstacles. Specifically, $\hat{p} = d(\{C^{t}_{obst}\}_{t=1}^T \!\sim\! g(p | c_c, c_g))$. The goal is to ensure the reconstructed motion plan $\hat{p}$ is the same as the given plan $p$, indicating $p$ is also optimal for the sampled obstacles $\{ C^t_{obst}\}_{t=1}^T$.

Based on a dataset $P$ of past motion plans, either from static/dynamic obstacle environments or completely open spaces, our Dyna-LfLH encoder and decoder find the optimal parameters $\psi^*$ for $g_\psi(\cdot)$ by minimizing a self-supervised loss 
\begin{equation}
\psi^* = \argmin_\psi \mathop{\mathbb{E}}_{\substack{p \sim P\\\hat{p}~=~d(\{C^t_{obst}\}_{t=1}^T \sim g_\psi(p|c_c, c_g))}} [\ell(p, \hat{p})],
\label{eqn::psi}
\end{equation}

where $\ell(\cdot, \cdot)$ is the reconstruction loss function to encourage the decoder output $d(\{C^t_{obst}\}_{t=1}^T)$ to be similar to the existing motion plan $p$.

\subsection{Dyna-LfLH}
By sampling $x^0_i$, $y^0_i$, $v^x_i$, and $v^y_i$ for all $N$ dynamic obstacles from $g_{\psi^*}$ and constructing $\{C^t_{obst}\}_{t=1}^T$ $K$ times, we can generate a supervised training set for Imitation Learning
\[
\mathcal{D}_{train} = \{(\{\{C^t_{obst}\}_{t=1}^T\}^k, p^k, c_c^k, c_g^k)\}_{k=1}^K, 
\]
with $K$ data points, where $p^k$ is (close to) optimal for $\{\{C^t_{obst}\}_{t=1}^T\}^k$. 
$\{\{C^t_{obst}\}_{t=1}^T\}^k$ can be transformed into observations in the form of LiDAR scans (with ray tracing) or depth images (with rendering).
With the training set $\mathcal{D}_{train}$, a parameterized motion planner $f_\theta(\cdot)$ can be learned by minimizing a supervised learning loss using gradient descent: 
\begin{equation}
\small
\theta^* = \argmin_\theta \mathop{\mathbb{E}}_{\substack{(\{C^t_{obst}\}_{t=1}^T, p, c_c, c_g)\\ \sim \mathcal{D}_{train}}} \big[ \ell(p, f_\theta(\{C^t_{obst}\}_{t=1}^T | c_c, c_g))\big].
\label{eqn::theta}
\end{equation}
$f_\theta$ will be used to produce motion plans based on the perceived $\{C^t_{obst}\}_{t=1}^T$ during deployment. Notice that during deployment, $\{C^t_{obst}\}_{t=1}^T$ is usually unavailable (unless using explicit future obstacle motion predictors). Therefore, we use the available history $\{C^t_{obst}\}_{t=-L+1}^0$ of length $L$ as inputs to $f_\theta$. Implementation details can be found below. 

\subsection{Implementation}
As shown in Algorithm~\ref{alg::dynalflh} line 2, a motion plan $p \in P \subset \mathcal{P}$ consists of a sequence of configurations and linear and angular velocities $\{(x^t, y^t, \textrm{yaw}^t), (v^t, \omega^t)\}_{t=1}^{T}$. We instantiate robot configurations in the robot frame so $c_c$ is always $\mathbf{0}$ and therefore ignored. $c_g$ is set to $(x^T, y^T, \textrm{yaw}^T)$, which is included in $p$ and ignored as well.  The encoder $g_{\psi}(\cdot)$ is a network of three one-dimensional convolutional layers followed by  fully connected autoregressive layers mapping the input motion plan $p$ to the distribution parameters of the dynamic obstacles' location and velocity (Eqn.~\ref{eqn::distribution}) in the form of means and log-variances. The decoder $d$ is a re-implementation of Ego-Planner \cite{zhou2020ego} with differential convex optimization layers \cite{agrawal2019differentiable}.
The reconstruction loss $\ell$ in Eqn. \ref{eqn::psi} is the mean squared error between all positions, linear and angular velocities $\{x^t, y^t, v^t, \omega^t\}_{t=1}^{T}$ in $p$ and their reconstructed values. 
To additionally regularize the loss $\ell$ in Eqn.~\eqref{eqn::psi}, we impose an obstacle location prior distribution to a normal distribution fitted on all positions $\{(x^t, y^t)\}_{t=1}^{T}$ in the plan $p$ as an additional loss $\ell_{prior}$ to prevent obstacles from getting too far away from the plan $p$. Similarly, an obstacle-obstacle and obstacle-plan collision regularization loss $\ell_{coll} = \sum\max(c-d,0)^2$ is added, where clearance $c=0.5$m and $d$ is the distance either between two obstacles or between the obstacle and its closest point on $p$. We use the same regularization weights as in LfLH~\cite{wang2021agile}.

To create $\mathcal{D}_{train}$, $S=4$ samples of a set of $N=1$ obstacles are taken from the latent space produced by $g_{\psi^*}(\cdot)$ for every motion plan $p$ (lines 5-14). 
$N$ dynamic obstacles, $\{\mathbf{C}^t_i\}_{i=1}^N,~ 1 \leq t \leq T$, which make motion plan $p$ optimal are constructed using Eqn.~\eqref{eqn::obstacle} and the observations are rendered as 2D LiDAR scans using ray casting given the current robot configuration $c^t$ along the plan $p$ and the corresponding obstacle configuration $C^t_{obst}$ (lines 8-9). 
$\mathcal{D}_{train}$ is augmented to enhance variability. First, up to five random non-colliding obstacles are added to increase variability in $C^t_{obst}$. Second, motion plans with velocities over 0.9 m/s are augmented without obstacles to help the model learn fast navigation in open spaces.

In our Dyna-LfLH implementation, $f_{\theta^*}(\cdot)$ (Eqn.~\eqref{eqn::theta}) learns to produce one single action $u^i$ given a sequence of $L=5$ historic LiDAR scans, $\{C^t_{obst}\}_{t=i-L+1}^i$, which comprise a data point in  $\mathcal{D}_{train}$ (lines 10-12). Each data point starts at the robot's current configuration $c^i_c = \mathbf{0}$ and the goal $c^i_g$ is a unit vector of a point 2.5 m away on the existing motion plan $p$. 

By taking in a history of $L$ LiDAR scans, the motion planner can implicitly encode and address obstacle dynamics. $f_{\theta^*}(\cdot)$ is modeled as a feed-forward recurrent neural network with two hidden layers, each of size 256 followed by a fully connected layer mapping the hidden layer to linear and angular velocities (line 16).
At each time step $m$ during deployment (lines 18-19), $c^m_c=\mathbf{0}$ and $c^m_g$ is instantiated as a unit vector of a point 2.5m away on the global path created using the \texttt{move\_base} stack. 

We use the same Model Predictive Control (MPC) model as LfLH \cite{xiao2021agile} to check for and avoid collisions, with the addition that future LiDAR scans are also simulated based on the two most recent scans. If a collision is imminent, the robot will stop and slowly reverse until no collision is predicted.

\begin{algorithm}
 \caption{Dyna-LfLH}
 \begin{algorithmic}[1]
 \renewcommand{\algorithmicrequire}{\textbf{Input:}}
 \REQUIRE existing motion plans $P$, obstacle number $N$,  sampling count $S$, history sequence length $L$
\\\hrulefill
  \STATE // \textbf{Learning Dynamic Hallucination}
  \STATE learn $\psi^*$ for $g_\psi(\cdot)$ with $P$ \hfill $\triangleright$ Eqn.\eqref{eqn::psi}
  \STATE // \textbf{Dataset Generation}
  \STATE $\mathcal{D}_{train} \leftarrow \emptyset$
  \FOR {every $p \in P$}
    \FOR {$S$ times}
        \STATE sample $\{(x^0_i, y^0_i), (v^x_i, v^y_i)\}_{i=1}^N$ with $g_{\psi^*}$ \hfill $\triangleright$ Eqn.~\ref{eqn::distribution}
        \STATE create $\{\mathbf{C}^t_i\}_{i=1}^N,~ 1 \leq t \leq T$ \hfill $\triangleright$ Eqn. \ref{eqn::obstacle} 
        \STATE render LiDAR scans for $\{C^t_{obst}\}_{t=1}^T$
        \FOR {every $u^i \in p, L \leq i \leq T-1$}
            \STATE $\mathcal{D}_{train}=\mathcal{D}_{train}\cup(\{C^t_{obst}\}_{t=i-L+1}^i, u^i, c^i_c, c^i_g)$
        \ENDFOR
    \ENDFOR
  \ENDFOR
  \STATE // \textbf{Dynamic Learning from Learned Hallucination}
  \STATE learn $\theta^*$ for $f_\theta(\cdot)$ with $\mathcal{D}_{train}$ \hfill $\triangleright$ Eqn. \ref{eqn::theta} 
\\\hrulefill
  \STATE // \textbf{Deployment} (each time step $m$)
  \STATE receive $\{C^t_{obst}\}_{t=m-L+1}^m, c^m_c, c^m_g$
  \STATE plan $p = u^m = f_{\theta^*}(\{C^t_{obst}\}_{t=m-L+1}^m~|~c^m_c, c^m_g)$
 \RETURN $p$
 \end{algorithmic}
 \label{alg::dynalflh}
 \end{algorithm}

\section{Experiments}
\label{sec::experiments}

\begin{figure*}
        \centering
      \includegraphics[width=2\columnwidth]{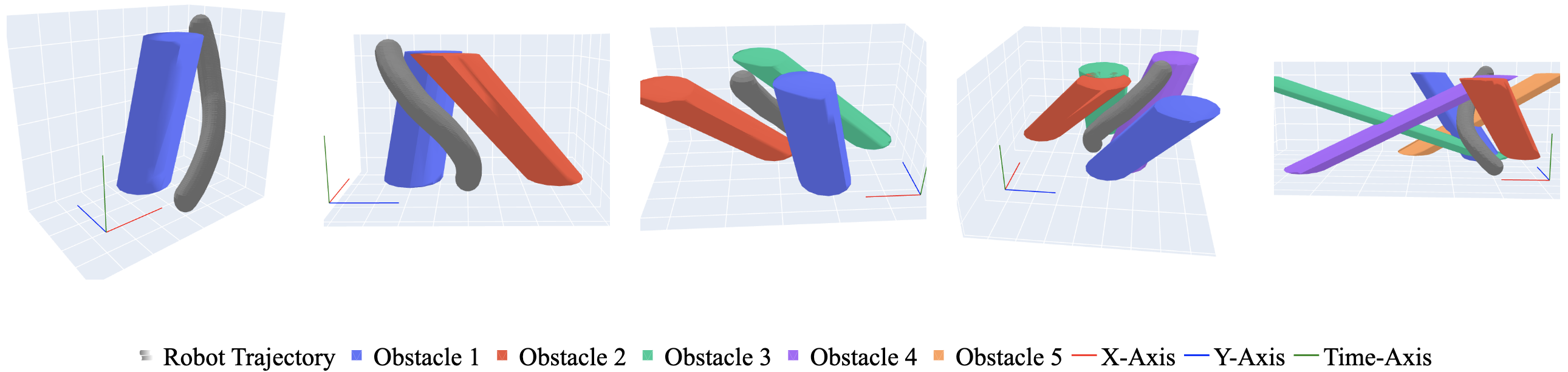}
  \caption{Example hallucinations of 1, 2, 3, 4, and 5 obstacles respectively. The Z-axis represents time, with the bottom indicating $t=1$ and the top indicating $t=T$. Robot trajectories are represented by dark gray while obstacles are colored. The robot and the obstacles start at the bottom and move to the top of the graph over time. The steepness of an obstacle is related with its speed. For example, obstacle 1 (blue) is moving slower than obstacle 2 (red) in the second 3D plot. In all cases, the robot trajectories maneuver through the obstacle(s) in a collision-free manner, while the obstacles are generated such that the robot trajectories are near-optimal.}
  \label{fig::hallucinations}
  \vspace{-10pt}
\end{figure*}

\begin{table}[htbp]
\centering
\caption{Key Parameters for DWA and Log‑MPPI}
\label{tab:params}
\begin{tabular}{@{}lcc@{}}
\toprule
\textbf{DWA}  & \textbf{Linear (x)} & \textbf{Angular}  \\
Max Velocity & 1.0 m/s & 1.57 rad/s \\
Min Velocity & 0.1 m/s & -1.57 rad/s \\
Acceleration Limit & 10.0 m/$\text{s}^2$  & 20.0 m/$\text{s}^2$  \\
Sampling Resolution & 12 & 40  \\
\midrule
Simulation Time & 2.0 s  \\
Simulation Granularity & 0.02 m \\
\midrule\midrule
\textbf{Log-MPPI}  & \textbf{Linear (x)} & \textbf{Angular}  \\
Max Velocity & 1.0 m/s & 1.5 rad/s \\
\midrule
Time Horizon & 6.0 s  \\
Sampling Rate & 50 $\text{s}^{\text{-1}}$  \\
Sampled Trajectories & 2496  \\
Exploration Variance & 1200.0  \\
State Dimension & 3  \\
Control Dimension & 2  \\
Cost Function Weights & [2.5, 2.5, 2]  \\
\bottomrule
\end{tabular}
\end{table}

Dyna-LfLH is implemented on a Clearpath Jackal robot, a four-wheeled, differential-drive, UGV, running the Robot Operating System \texttt{move\textunderscore base} navigation stack. 
The Jackal has a 720-dimensional, front-facing, 2D LiDAR with a 270$^\circ$ field of view, which is used to instantiate obstacle configuration $C^t_{obst}$. Dyna-LfLH is used as a local planner. 
We conduct both simulated and physical experiments to validate our hypothesis that Dyna-LfLH can learn to hallucinate dynamic obstacle configurations where previous motion plans are near-optimal, and agile motion planners can be learned through the learned hallucination.

\subsection{Baselines}
Dyna-LfLH is compared with a classical sampling-based motion planner~\cite{fox1997dynamic}, a model predictive path integral (MPPI) controller~\cite{log_mppi}, a state-of-the-art LfH approach~\cite{wang2021agile}, and an IL method~\cite{amir2023socialnav} trained on a large expert dataset~\cite{karnan2022socially}. Specifically, the Dynamic Window Approach (DWA)\cite{fox1997dynamic} samples actions and evaluates them with a cost function, and Log-MPPI\cite{log_mppi} samples from a log-normal mixture distribution and incorporates a 2D costmap. The key parameters for DWA and Log-MPPI are provided in Tab.~\ref{tab:params}.
LfLH is trained on 25129 data points (under 10 minutes) of a Jackal robot exploring an open space at 1.0 m/s. After training, it generates 10 static obstacles per data point. Behavior Cloning (BC)~\cite{amir2023socialnav} is a fully supervised approach using an 8.7-hour expert dataset in dense, dynamic spaces. Dyna-LfLH trains on the same open-space dataset as LfLH (under 10 minutes), generates $N=1$ dynamic obstacles, and learns a motion planner with different obstacle sequence lengths ($L=1, 3, 5, \text{and~} 10$ previous LiDAR scans). For consistency, all planners have a max linear velocity of 1.0 m/s, the maximum speed in the training data for LfLH and \dynalflh.

\subsection{Learned Dynamic Hallucination}
In Fig.~\ref{fig::hallucinations}, we present examples of the hallucination results. Five different latent spaces with one to five dynamic obstacles are learned. The corresponding obstacles are sampled from these learned distributions and visualized in a 3D space with the Z-axis representing time. The robot's trajectories successfully navigate through the obstacles without any collision at each time step, while the obstacles are essential to make the robot's maneuvers near-optimal, i.e., the obstacles are the reason why the robot needs to execute such an obstacle avoidance maneuver. Fig.~\ref{fig::hallucinations} also illustrates that, in most cases, hallucinating just one obstacle is sufficient in explaining the robot's movement, so hallucinating more than one obstacle is unnecessary. Furthermore, the figure shows that the hallucination function learns that both fast and slow-moving obstacles can be avoided using the same motion plan.

\subsection{Simulation}
\begin{figure}
  \centering
  \includegraphics[width=1.0\columnwidth]{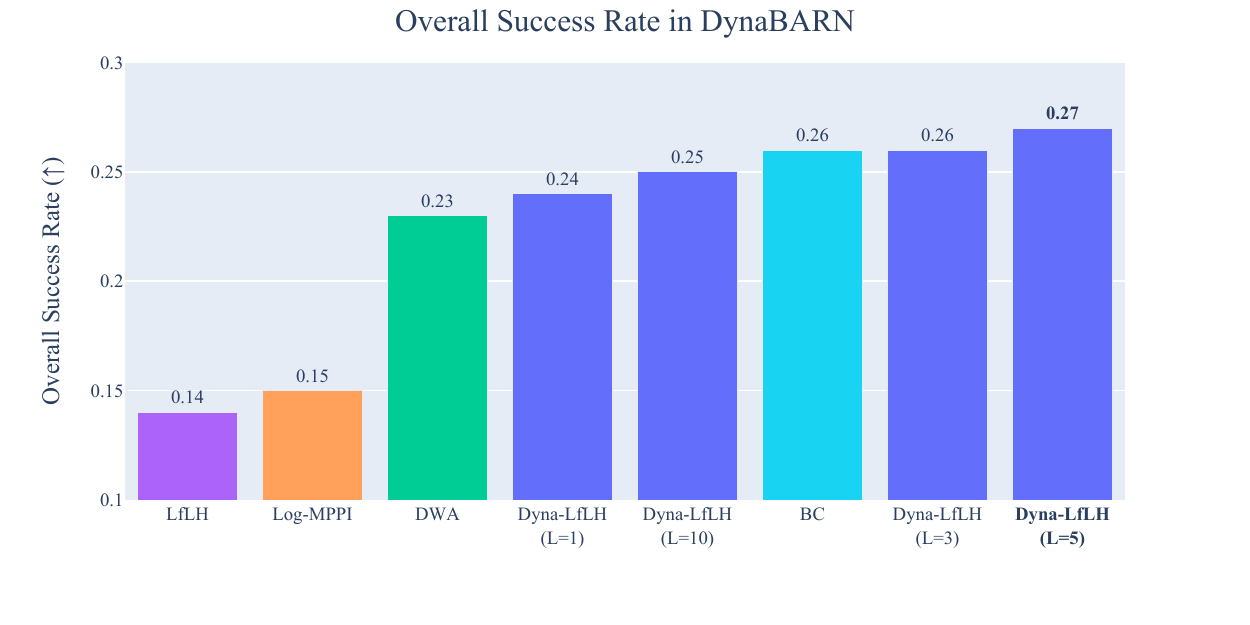}
  \caption{Simulation Results of 180 trials in DynaBARN. Dyna-LfLH with 5 history scans (L=5) performs the best overall. }
  \vspace{15pt}
  \label{fig::dynabarn_results}
  \vspace{-10pt}
\end{figure}

We use DynaBARN~\cite{nair2022dynabarn}, a simulation testbed for evaluating dynamic obstacle avoidance with a diverse set of 60 environments at different difficulty levels based on the number of obstacles and obstacle motion profiles.
For each DynaBARN environment, the robot navigates from one side to the other, and a single collision with any of the obstacles counts as a failure. 
For every method (DWA, Log-MPPI, LfLH, BC, and Dyna-LfLH), we run 3 navigation trials for each of the 60 environments for a total of 180 trials. In the simulation experiments, the recovery behaviors of all models are turned off to focus only on the main planner's performance. 
The overall results in terms of success rate in DynaBARN are shown in Fig.~\ref{fig::dynabarn_results}. 

The results show that LfLH does not perform well in DynaBARN due to a lack of consideration of obstacle dynamics during hallucination. The state-of-the-art classical planner, Log-MPPI~\cite{log_mppi}, also performs very poorly in this fast-moving and highly cluttered benchmark. DWA performs significantly better than LfLH and Log-MPPI. BC achieves good performance in the simulated DynaBARN after learning on 8.7-hours of expert demonstration data. Dyna-LfLH, trained on 10-minute self-supervised exploration in an open space, achieves a comparable success rate. The Dyna-LfLH with 5 history scans outperforms BC and achieves the best performance across the board. 

\subsection{Physical Experiments}

We also compare the best Dyna-LfLH planner ($L=5$) with the best planner in each other category, classical (DWA) and IL (BC), in a physical test course, 20 trials each (Fig. \ref{fig::dynaInAction}), with a total of 60 physical trials. We create an enclosed $8.2~\text{m}\times4.3~\text{m}$ arena with 11 randomly moving obstacles (iRobot Roombas of 0.33 m diameter). The Roombas move with a maximum linear speed of 0.5 m/s with a combination of behaviors such as spiraling outwards, following walls, and bouncing off each other. The success rate and average traversal time on success (TTS) are shown in Tab. \ref{tab::physical_results}. 

Dyna-LfLH achieves the best success rate up to 50\%, a 25\% improvement over the 2nd best planner, DWA. BC does not work well in the real world and only achieves 15\%. We also report the traversal time for each method and Dyna-LfLH achieves the fastest navigation among all successful trials. 
For safety in the physical experiments, the recovery behaviors of all three methods are turned on.

\begin{table}
\centering
\scriptsize
\caption{Physical Experiment Results.}
\begin{tabular}{cccc}
\toprule
Metrics & DWA  & BC & Dyna-LfLH\\ 
\midrule
Success Rate ($\uparrow$)  & 0.40 & 0.15 & \textbf{0.50}\\
Avg. TTS (s, $\downarrow$) & 17.26~$\pm$~8.31 & 17.67~$\pm$~2.52 & \textbf{12.60}~$\pm$~\textbf{2.37}\\
\bottomrule
\end{tabular}
\label{tab::physical_results}
\end{table}

\subsection{Discussions}
Navigating dynamic environments poses a significant challenge for autonomous robots, as evidenced by the low success rates of all evaluated methods (Fig.~\ref{fig::dynabarn_results}). The simulated DynaBARN environments present particularly difficult scenarios due to the high variability in obstacle motion profiles — individual obstacles frequently change speed and direction, often moving faster than the robot's maximum speed and making sudden, sharp turns. In contrast, the Roombas used in the physical experiments move generally at a constant velocity of 0.5 m/s, which is slower than the robot's maximum speed. This difference in obstacle dynamics makes the physical experiments comparatively easier, contributing to the higher success rates observed in real-world trials.

Nonetheless, our visualization of the learned dynamic obstacles validate our hypothesis that Dyna-LfLH successfully learns to generate dynamic obstacles for which our existing motion plans are optimal. The results from the simulation and physical experiments show that a dynamic motion planner can be learned from such hallucinated data. The learned motion planner also demonstrates generalization to both simulation and the real world. 

However, we observe that while theoretically possible to approximate $\{\{C^{t, i}_{obst}\}_{t=1}^T\}_{i=1}^\infty$ using a learned distribution, in practice, distributions in Eqn.~\eqref{eqn::hallucination} suffer from mode collapse, producing limited samples that fail to approximate the infinitely many obstacle configuration sequences, $\{\{C^{t, i}_{obst}\}_{t=1}^T\}_{i=1}^\infty$. This likely contributes to the low success rate in Fig.~\ref{fig::dynabarn_results}. A potential solution is to change the architecture and gradually add variance to the hallucinated obstacles.

\section{Conclusions}
\label{sec::conclusions}
Dyna-LfLH is a self-supervised method for mobile robot navigation in dynamic environments, capable of navigating among fast-moving, unpredictable obstacles using only past deployment data or data from open spaces. 
It generates dynamic obstacle configurations to optimize motion plans and provide efficient training data for planners. 
Both simulated and physical experiments show that a local planner trained with Dyna-LfLH outperforms classical planning methods and supervised approaches that rely on large expert datasets.

\bibliographystyle{IEEEtran}
\bibliography{IEEEabrv,references}

\end{document}